%% file: root.tex
\documentclass{article}
\usepackage{iclr2027_conference,times}
\input{math_commands.tex}

\usepackage{amsmath,amssymb,amsfonts}
\usepackage{booktabs}
\usepackage{multirow}
\usepackage{graphicx}
\usepackage{url}
\usepackage{xcolor}
\usepackage{placeins}
\usepackage{float}
\usepackage{capt-of}
\usepackage{hyperref}
\newcommand{\method}{G$^2$WAM}
\newcommand{\secondbest}[1]{\underline{#1}}

\title{G$^2$WAM: World Action Models for Efficient and Geometry Grounded Image-Goal Navigation}

\author{
\parbox{0.95\textwidth}{\centering
Ke Gui\textsuperscript{1},
Chenghao Shi\textsuperscript{1},
Ziyue Wang\textsuperscript{1},\\
Longwen Chen\textsuperscript{1},
Xieyuanli Chen\textsuperscript{1},
Huimin Lu\textsuperscript{1,2,3}
\\[0.6em]
{\normalfont\small
\textsuperscript{1}College of Intelligence Science and Technology,\\
National University of Defense Technology, China
\\[0.3em]
\textsuperscript{2}National Key Laboratory of Equipment State Sensing
and Smart Support,\\
National University of Defense Technology, China
\\[0.3em]
\textsuperscript{3}School of Future Technology,\\
National University of Defense Technology, China
}
}
}

\iclrfinalcopy

\begin{document}
\maketitle
\lhead{Preprint}

\begin{abstract}
World models capture environment dynamics and support counterfactual reasoning for planning. In image-goal navigation, where the target often lies beyond the current field of view, the agent must not only model environmental dynamics but also maintain a persistent spatial memory of the scene. Existing approaches either introduce spatial context by fine-tuning pretrained models on the target task, which underexploits their spatial priors, or rely on external 3D reconstruction, which is computationally expensive for real-world deployment. In this paper, we propose the Geometry Grounded World Action Model (\method{}) for image-goal navigation. \method{} encodes spatial cues through a learned scene-coordinate localization module that anchors the current, goal, and history frames in a common reference frame, yielding a stable and consistent spatial context that enables efficient navigation without explicit 3D reconstruction. Following the world-action-model paradigm, \method{} jointly predicts future states and actions, eliminating the need for costly cross-entropy method (CEM) style planning. In addition, \method{} operates entirely in latent space, avoiding the computational burden of pixel-level reconstruction and substantially reducing inference cost. Across four trajectory prediction benchmarks, \method{} reduces average ATE and RPE by 40.1\% and 35.2\%, respectively. In closed-loop navigation experiments, it improves the success rate from 54.0\% to 63.0\% and the success weighted by path length~(SPL) from 36.4\% to 59.2\%, while maintaining high efficiency.
\end{abstract}

\section{Introduction}
\label{sec:intro}
Image-goal navigation asks an agent to reach the location shown in a target image by using its stream of first-person observations. Recent advances have progressed from reactive end-to-end policies~\citep{gnm,nomad,vint2023} to world-model-based approaches~\citep{nwm,dinowm,relnwm,dreamerv3}, in which the world model serves as a learned model of environment dynamics. Concretely, it can jointly capture navigation priors and scene memory within a single training pipeline, and support counterfactual reasoning that helps the agent solve complex long-horizon navigation tasks, including those whose targets lie beyond the current field of view. In practice, however, error accumulation over multi-step rollouts causes long-horizon imagination to degrade rapidly~\citep{nwm,drnwm}. Moreover, cross-entropy method (CEM) planning requires repeated forward evaluations, incurring high inference cost and latency, and the selected open-loop trajectory must still be converted into closed-loop waypoints at execution time, which introduces additional error.  

Existing methods attempt to explicitly inject depth information during training~\citep{geometryforcing}, or to bolt on external 3D reconstruction~\citep{gen3c,evoworld}, Gaussian-splatting~\citep{ma2026_4dgswam}, or latent spatial memory~\citep{spmem,worldpack,mirage}, so as to provide the world model with explicit spatial context that improves long-horizon and spatial consistency. Yet these methods either exploit spatial priors insufficiently or are too heavy for practical deployment. Our key insight is that, when navigating, humans do not maintain a precise geometric model of the environment~\citep{warren2019noneuclidean,ishikawa2006spatial}; instead, they rely on an egocentric, reactive representation~\citep{wang2002human}, which can be captured by the relative pose between the current view and the goal image.

\begin{figure}[t]
\centering
\includegraphics[width=0.98\textwidth]{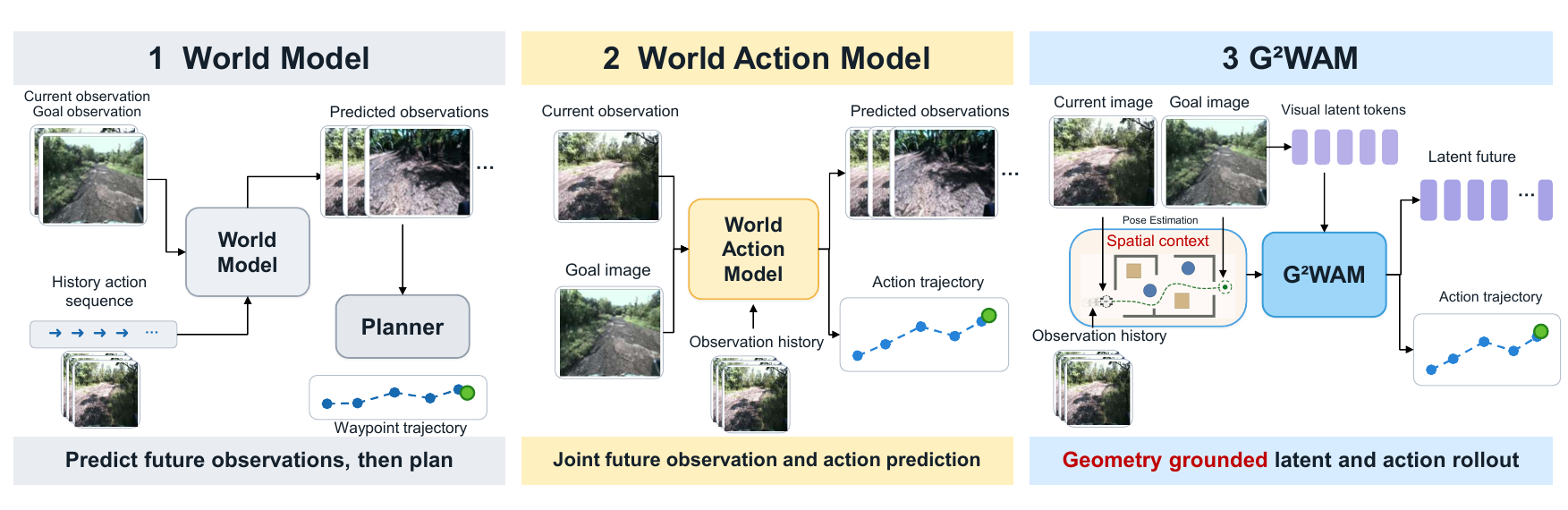}
\caption{Navigation world model paradigms. Conventional world models plan over predicted observations, whereas world action models jointly predict observations and actions. \method{} further leverages estimated spatial context to efficiently roll out future latents and executable actions.}
\label{fig:motivation}
\end{figure}

We therefore present the Geometry Grounded World Action Model (\method{}) for image-goal navigation. \method{} encodes environmental information into an end-to-end localization network and provides the agent with egocentric relative pose to the goal as well as the history trajectory as spatial cues. By anchoring all observation frames and the goal frame to fixed positions in space, we obtain a stable, geometrically consistent spatial context.  In designing \method{}, we paid particular attention to its efficiency for practical deployment. To this end, \method{} adopts the world-action-model paradigm to jointly predict action chunks and latent future states, sidestepping the expensive CEM-style planning. It further keeps the entire rollout in latent space, avoiding pixel-level future reconstruction. Fig.~\ref{fig:motivation} depicts the key difference between our method and existing methods based on world models and world action models.

Our contributions can be summarized as follows:
\begin{itemize}
  \item We propose \method{}, a novel implicit world action model for navigation that jointly predicts latent future states and executable action chunks, avoiding the costly CEM-style planning and heavy pixel-level reconstruction.

  \item We introduce a new paradigm for spatial context aiding. Instead of reconstructing explicit geometry, we anchor the observation and goal frames in a shared end-to-end localization network, which efficiently provides a continuous, geometrically consistent spatial context over the navigation history and the relative target.

  \item Extensive experiments demonstrate that \method{} achieves the best overall trajectory prediction and navigation performance, reducing average ATE and RPE by 40.1\% and 35.2\% in trajectory prediction, and improving the success rate from 54\% to 63\% (a 9-point gain) and success weighted by path length (SPL) from 36.4\% to 59.2\% (a 22.8-point gain) in closed-loop navigation, while maintaining high efficiency. We will open-source our approach to benefit the research community.
\end{itemize}

\section{Related Work}
\label{sec:related}

\subsection{Visual Navigation}

Experience-based navigation uses recorded routes, demonstrations, or
topological graphs to guide motion~\citep{furgale2010vtr,pathak2018zeroshot,sptm,ving}.
Goal-conditioned policies instead map current and goal images directly to
actions; representative methods include GNM~\citep{gnm}, ViNT~\citep{vint2023},
and NoMaD~\citep{nomad}. These methods do not explicitly predict future world
states.
Metric goal estimation provides complementary geometric guidance. PointGoal
methods use localization or visual odometry~\citep{ddppo,partsey2022mapping},
while image-based localization estimates pose through feature matching or
scene-coordinate regression~\citep{superglue,dsac,rscore}.
AnyImageNav~\citep{anyimagenav2026} uses recovered goal pose for precise
navigation. 
Reactive policies have achieved promising results, but they lack modeling of environment dynamics, making them prone to pattern matching. Accordingly, recent state-of-the-art methods have progressed toward world-model-based approaches.

% In our known-scene setting, \method{} similarly estimates the
% current-to-goal relative pose, but uses it to condition latent dynamics and
% action prediction.

\subsection{World Models and World Action Models for Navigation}

Navigation world models predict the consequences of actions in image or latent
space. NWM~\citep{nwm} generates future images, whereas DINO-WM~\citep{dinowm},
LS-NWM~\citep{lsnwm}, and ReL-NWM~\citep{relnwm} support planning with predicted
visual features or latent states. World action models further couple state and
action prediction. Spatial-perceiving WAM~\citep{swam2026} jointly predicts
RGB-D observations and waypoints, NavWAM~\citep{navwam2026} models observations,
goal progress, and action chunks, and WAMNav~\citep{wamnav2026} jointly predicts
latent dynamics and actions. \method{} also performs joint latent state-action
prediction, but introduces explicit spatial context through relative pose and
causal motion conditioning. Unlike existing WAMs, it predicts metric waypoints
through an implicit latent rollout without reconstructing future observations.

Geometry-aware world models mainly improve spatial consistency in two ways. The first direction maintains geometry-grounded spatial memory. GEN3C~\citep{gen3c} and EvoWorld~\citep{evoworld} construct explicit 3D scene presentations and reproject them into target views to condition visual generation. Other methods improve long-term consistency by organizing historical information according to scene geometry: long-term spatial memory~\citep{spmem} retrieves previously observed scene content, WorldPack~\citep{worldpack} selects and compresses observations using camera pose and field-of-view overlap, and Mirage~\citep{mirage} stores depth-unprojected latent tokens in a persistent 3D cache. The second direction introduces geometric constraints without maintaining a persistent scene representation. Geometry Forcing~\citep{geometryforcing} aligns intermediate generative features with geometric representations during training, while DR-NWM~\citep{drnwm} uses future anchors and bidirectional epipolar guidance to reduce geometric drift during visual rollout. 

In contrast, \method{} neither reconstructs an explicit 3D scene nor generates future images under geometric constraints. Instead, it derives spatial context from estimated poses, a lightweight metric condition for jointly rolling out latent states and executable actions.

\section{Method}
\label{sec:method}
\begin{figure}[!t]
\centering
\includegraphics[width=0.98\textwidth]{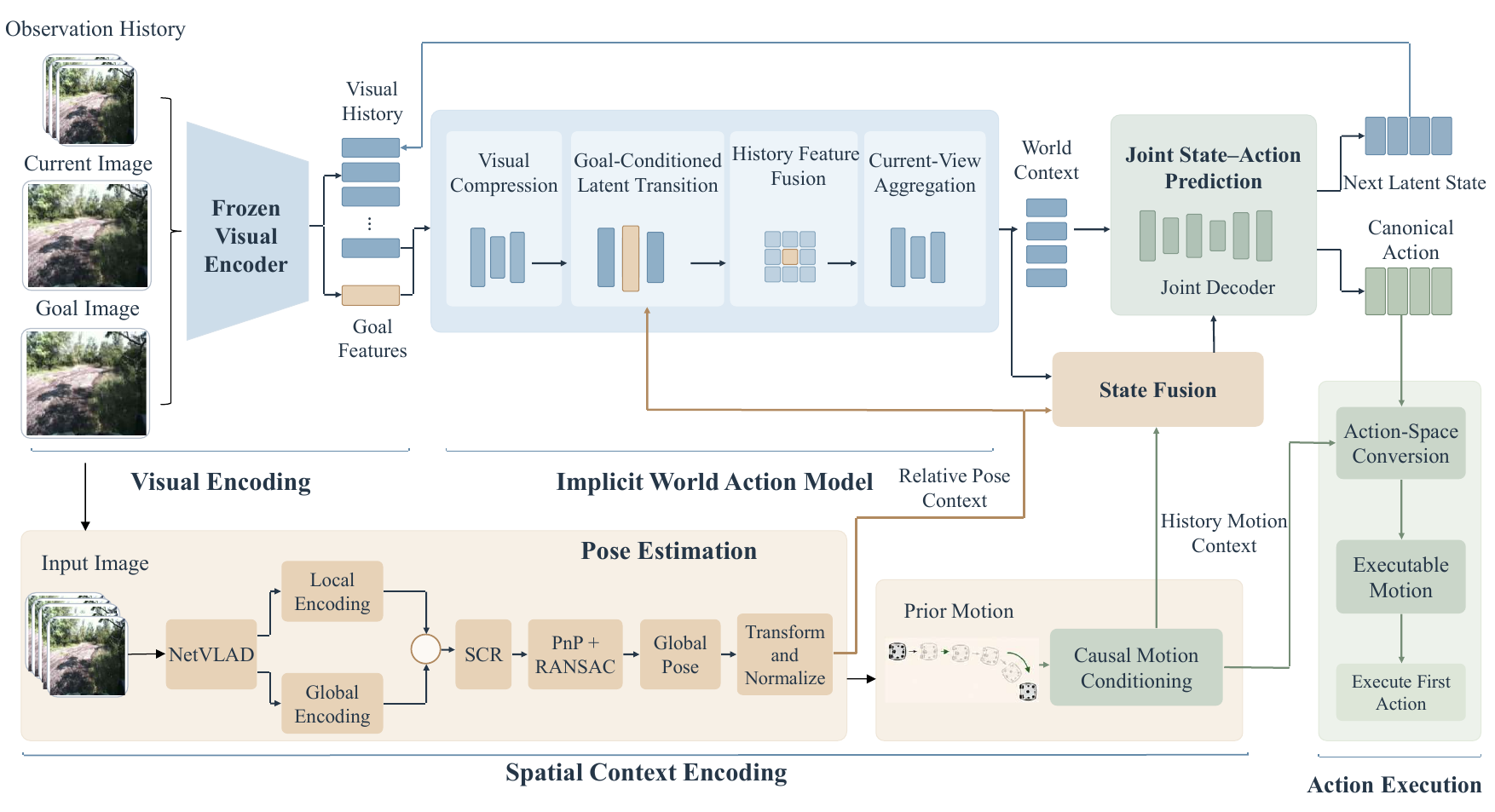}
\caption{Overview of \method{}. A learned localizer anchors the goal and observed frames in a shared scene frame to obtain the goal-relative pose and causal history motions. The goal image and pose guide joint latent and waypoint prediction, while the history motions condition the action branch.}
\label{fig:arch}
\end{figure}

We build \method{} around a single design principle: efficient, geometry-grounded navigation. This principle is realized through two design choices. First, the spatial context is geometry-grounded without costly 3D reconstruction. A lightweight scene-coordinate localization module anchors the goal and observed frames in a shared scene frame, so the relative pose between the current view and the goal, together with the history motions, provides an egocentric geometric context that grounds the rollout in the target. Second, the model is efficient by construction. It adopts the world-action-model paradigm and rolls out entirely in latent space, avoiding CEM-style planning and pixel-level reconstruction. Together, an implicit world action model consumes this lightweight relative-geometry context and rolls it forward, jointly predicting latent future states and canonical waypoint increments. The overall architecture is shown in Fig.~\ref{fig:arch}.

We first describe the spatial context induced by scene-coordinate anchoring (Sec.~\ref{sec:spatial_context}), then the implicit world action model (Sec.~\ref{sec:implicit_wam}), and finally how relative-pose and motion conditioning adapt the model to a concrete platform (Sec.~\ref{sec:conditioning}). Training is described in Sec.~\ref{sec:training}.

% \vspace{-20pt}

\subsection{Spatial Context via Scene-Coordinate Anchoring}
\label{sec:spatial_context}
Our key insight is that an agent need not build an explicit geometric model of the environment to navigate. It needs a stable, geometrically consistent spatial context indicating where the goal lies relative to its history of visited locations. We obtain this by anchoring the observation history and the goal image in a shared scene coordinate through an end-to-end localization network.

Concretely, the localization module estimates the current, goal, and history-frame poses in a shared scene frame, from which we compose the current-to-goal relative transformation and the consecutive motion increments. Let $\bm I_j$ be the image at index $j$, and let $\hat{\bm T}_j^{\mathrm S}\in\mathrm{SE}(3)$ map its calibrated robot frame into the shared scene frame $\mathrm S$, where $t$, $g$, and $i$ index the current, goal, and history frames. We first obtain each pose through a scene-specific image localization module $f_{\mathrm{loc}}$:
\begin{equation}
\label{eq:image_to_pose}
\hat{\bm T}_j^{\mathrm S}=f_{\mathrm{loc}}(\bm I_j;\mathcal D_{\mathrm S}),
\qquad j\in\{t-3,\ldots,t,g\},
\end{equation}
where $\mathcal D_{\mathrm S}$ denotes pre-collected scene data and camera-to-robot calibration is included in the interface. We then compute
\begin{equation}
\label{eq:relative_transforms}
\hat{\bm T}_{t\to g}
=(\hat{\bm T}_t^{\mathrm S})^{-1}\hat{\bm T}_g^{\mathrm S},
\qquad
\hat{\bm T}_{i\to i+1}
=(\hat{\bm T}_i^{\mathrm S})^{-1}\hat{\bm T}_{i+1}^{\mathrm S},
\quad i=t-3,\ldots,t-1.
\end{equation}
The planar translation of $\hat{\bm T}_{t\to g}$ is converted to the model's coordinate convention and divided by the waypoint spacing $\ell$ to form $\bm p_t$; the consecutive transforms give the history-dependent entries of $\bm m_t$. These are embedded as $\bm e_t^{\mathrm{pose}}$ and $\bm e_t^{\mathrm{motion}}$ (Sec.~\ref{sec:conditioning}) and condition the latent rollout in Sec.~\ref{sec:implicit_wam}. This design separates the world action model from the localization network: since the model consumes only the current-to-goal relative transformation and recent causal motion, the localization network can be any end-to-end method or a traditional place-recognition-plus-pose-estimation pipeline, without changing the architecture. This plug-and-play property makes the world action model agnostic to the localization interface. By default, we use R-SCoRe~\citep{rscore}, a state-of-the-art image-based end-to-end localizer that maps images to geometrically grounded correspondences for robust 6-DoF pose estimation, achieving 64.3\% success within $(0.25\,\mathrm{m},2^\circ)$ at 140--270\,ms per query on an RTX 2080.

Note that scene-coordinate anchoring assumes a previously seen scene, as it requires pre-collected trajectory data. This is practically meaningful because our method targets out-of-field-of-view navigation. In a completely unknown scene, such a task can rely only on a world model's imagination, which is empirical and unreliable since any configuration is possible. Existing methods therefore resort to fine-tuning or external 3D reconstruction. Our method instead anchors observations and the goal to fixed spatial positions via end-to-end localization, exploiting scene information more directly than fine-tuning and more lightly than 3D reconstruction.

\subsection{Implicit World Action Model}
\label{sec:implicit_wam}
The world action model is built to roll out task-relevant state without ever reconstructing an image. This choice is deliberate: navigation only requires the model to track where the agent is heading, not to reproduce the appearance of the scene, so an abstract latent rollout suffices. We therefore predict the next patch latent and the corresponding action from a shared goal-conditioned transition context, never passing through an image decoder. The relative pose $\bm p_t$ and motion profile $\bm m_t$ from Sec.~\ref{sec:spatial_context} enter this model through their embeddings $\bm e_t^{\mathrm{pose}}$ and $\bm e_t^{\mathrm{motion}}$, defined in Sec.~\ref{sec:conditioning}.

A frozen DINOv3 encoder~\citep{dinov3} maps the observation history and goal image into patch latents $\bm z_{t-H+1:t}\in\mathbb{R}^{H\times P\times D}$ and $\bm z_g\in\mathbb{R}^{P\times D}$, where $H=4$ is the context length, $P$ the number of patches, and $D=384$ the latent dimension. The goal embedding is $\bm q_g=\phi_g(\operatorname{pool}(\operatorname{LN}(\bm z_g)))$,
where $\phi_g$ is a two-layer MLP, $\operatorname{LN}$ is layer normalization, and $\operatorname{pool}$ denotes mean pooling over patches.
At rollout step $k\in\{0,\ldots,7\}$, the model maintains a fixed-length latent history and computes

\begin{equation}
\label{eq:implicit}
\begin{aligned}
\bm c_{t+k}
&=\operatorname{Transition}\!\left(
\bm z_{t+k-H+1:t+k},\bm q_g,\bm e_t^{\mathrm{pose}}
\right),\\
\bm s_{t+k}
&=\operatorname{pool}(\bm c_{t+k})
+\bm e_t^{\mathrm{pose}}+\bm e_t^{\mathrm{motion}},\\
\left(\bm z^{\mathcal J}_{t+k+1},
\hat{\bm a}_{t+k}^{\mathrm{can}}\right)
&=\mathcal J\!\left(\bm c_{t+k},\bm s_{t+k}\right),\\
\hat{\bm z}_{t+k+1}
&=\operatorname{LN}\!\left(
\alpha_{\mathcal J}\bm z^{\mathcal J}_{t+k+1}
+\alpha_{\mathrm B}\phi_{\mathrm B}(\bm c_{t+k})\right).
\end{aligned}
\end{equation}

The transition module fuses the latent history with the goal and relative-pose
conditions to produce the patch-level context
$\bm c_{t+k}\in\mathbb{R}^{P\times D}$. Its pooled representation is combined
with the pose and motion embeddings to obtain the compact step state
$\bm s_{t+k}\in\mathbb{R}^{D}$. The joint decoder $\mathcal J$ then predicts
the next latent candidate and the canonical waypoint increment
$\hat{\bm a}^{\mathrm{can}}_{t+k}=(\Delta x,\Delta y,\Delta\theta)$.
As defined in Eq.~\ref{eq:implicit}, the latent candidate is combined with a
base transition and appended to the fixed-length history, yielding an
autoregressive latent rollout without image decoding. Detailed transition,
conditioning, and decoder architectures are provided in
Appendix~\ref{app:implementation}.

\subsection{Spatial Context Conditioning And Action Adaption}
\label{sec:conditioning}
The two conditioning signals play distinct roles: the relative pose supplies target geometry, while the motion profile records the past trajectory and specifies how the canonical prediction is converted into executable motion.

The relative-pose conditioner receives
$\bm p_t=(\Delta x_t^g,\Delta y_t^g,
\sin\Delta\theta_t^g,\cos\Delta\theta_t^g)$.
This four-dimensional vector is derived from $\hat{\bm T}_{t\to g}$ in Eq.~\ref{eq:relative_transforms}.
Here $\Delta x_t^g$ and $\Delta y_t^g$ are the current-to-goal planar
translation in the model's coordinate convention and in waypoint
units, while $\Delta\theta_t^g$ is the relative yaw. The conditioner
separates translation from rotation and computes
\begin{equation}
\label{eq:pose}
\bm e_t^{\mathrm{pose}}
=
\phi_{\mathrm{fuse}}\!\left(
\left[
\phi_{\mathrm{trans}}\!\left(
\bm u_t,\log(1+d_t)
\right),
\phi_{\mathrm{yaw}}(\bm v_t)
\right]
\right),
\end{equation}
where
$\bm u_t=(\Delta x_t^g,\Delta y_t^g)/\max(d_t,\epsilon)$ is the
normalized goal-direction vector,
$d_t=\sqrt{(\Delta x_t^g)^2+(\Delta y_t^g)^2}$ is the goal distance in
waypoint units,
$\bm v_t=(\sin\Delta\theta_t^g,\cos\Delta\theta_t^g)$ is the
relative-yaw vector, and $\epsilon>0$ prevents division by zero.
The brackets denote concatenation. The functions
$\phi_{\mathrm{trans}}$ and $\phi_{\mathrm{yaw}}$ are separate
two-layer MLPs. The fusion function $\phi_{\mathrm{fuse}}$ applies a
two-layer MLP followed by layer normalization.

Independently, the motion conditioner receives the five-dimensional profile formed from the consecutive transforms in Eq.~\ref{eq:relative_transforms},
$\bm m_t=[\bar d_t,\bar\theta_t,d_s,\theta_s,\ell]^{\mathsf T}$.
The history-dependent terms are the median per-step distance
$\bar d_t$, divided by waypoint spacing $\ell$, and the mean
absolute per-step yaw change $\bar\theta_t$. The dataset-level
translation and yaw scales are $d_s$ and $\theta_s$, respectively;
$\ell$ is the waypoint spacing in metres. A two-layer MLP with layer
normalization maps this profile to the motion embedding
$\bm e_t^{\mathrm{motion}}
=\phi_{\mathrm{motion}}(\log(\bm 1+\bm m_t))$,
where the logarithm is applied componentwise.

A canonical-motion adapter $A$ maps each canonical prediction to
the action interface:
\begin{equation}
\label{eq:motion_adapter}
\hat{\bm a}_{t+k}
=
A\!\left(\hat{\bm a}_{t+k}^{\mathrm{can}},\bm m_t\right)
=
\left[
\mathcal N_{xy}\!\left(
\tilde d_t\hat{\bm a}_{t+k,xy}^{\mathrm{can}}
\right);
\tilde\theta_s\hat a_{\theta,t+k}^{\mathrm{can}}
\right].
\end{equation}
Here $\hat{\bm a}_{t+k,xy}^{\mathrm{can}}$ denotes the two-dimensional canonical translation, $\hat a_{\theta,t+k}^{\mathrm{can}}$ the scalar canonical yaw, $\tilde d_t=\operatorname{clip}(\bar d_t,0.1,4.0)$, and $\tilde\theta_s=\operatorname{clip}(\theta_s,0.05,1.0)$. The semicolon denotes concatenation. For a two-dimensional input $\bm\xi$, the planar normalization is $\mathcal N_{xy}(\bm\xi)=2(\bm\xi-\bm b_{\min})\oslash(\bm b_{\max}-\bm b_{\min})-\bm 1$, where $\bm b_{\min}$ and $\bm b_{\max}$ are fixed two-dimensional action bounds, $\oslash$ denotes componentwise division, and $\bm 1$ is the two-dimensional all-ones vector. This function maps translation in waypoint units to the normalized planar action interface. Thus $\bm p_t$ supplies target geometry, whereas $\bm m_t$ specifies how the canonical prediction is converted into platform-specific motion.

\subsection{Training}
\label{sec:training}

We train the \method{} end-to-end with a multi-term objective,
while the localization interface is trained separately. The latent loss
combines a cosine term and a small mean-squared-error term, the action loss is a
smooth-$\ell_1$ on the canonical action, and trajectory supervision compares
cumulative waypoint coordinates and yaw over the rollout. Rare turns receive
extra weight so the model does not collapse to always-advance behavior:

\begin{equation}
\label{eq:loss}
\mathcal{L} = \lambda_{z}\mathcal{L}_{z} + \lambda_{a}\mathcal{L}_{a}
+ \lambda_{\mathrm{traj}}\mathcal{L}_{\mathrm{traj}}
+ \lambda_{\mathrm{yaw}}\mathcal{L}_{\mathrm{yaw}},
\end{equation}

The two direct prediction terms are
\begin{equation}
\label{eq:direct_losses}
\begin{aligned}
\mathcal L_z &= 1-\operatorname{Cos}(\hat{\bm Z},\bm Z^*)
                  +0.3\operatorname{MSE}(\hat{\bm Z},\bm Z^*),\\
\mathcal L_a &=
\frac{\sum_{b,k}w_{b,k}\,\ell_{\mathrm{SL1}}
 (\hat{\bm a}^{\mathrm{can}}_{b,t+k},\bm a^{\mathrm{can},*}_{b,t+k})}
 {\sum_{b,k}w_{b,k}},
\end{aligned}
\end{equation}
where $\hat{\bm Z}$ and $\bm Z^*$ stack the predicted and target patch latents across rollout steps, $\operatorname{Cos}$ averages patchwise cosine similarity, $\operatorname{MSE}$ averages squared feature error, and $\ell_{\mathrm{SL1}}$ averages smooth-$\ell_1$ error over the three canonical-action components. The weight $w_{b,k}$ increases for turning steps according to the target executable yaw increment and applies to the entire action increment. The remaining terms $\mathcal L_{\mathrm{traj}}$ and $\mathcal L_{\mathrm{yaw}}$ compare cumulative executable waypoint positions and yaw using $\ell_1$ losses, with planar increments unnormalized before accumulation. The scalar coefficients $\lambda_z,\lambda_a,\lambda_{\mathrm{traj}},\lambda_{\mathrm{yaw}}$ weight these four losses, whose values are $0.3$, $3.0$, $2.0$, and $0.5$, respectively. We use AdamW with batch size $32$, zero weight decay, and gradient-norm clipping at $10$.

\section{Experiments}
\label{sec:experiments}
We conduct our experiments to show that \method{} (i) achieves state-of-the-art trajectory prediction and, with a stronger localization method, holds sufficiently high potential for even better prediction; (ii) retains performance comparable to the baselines even when transferred zero-shot to a new scene where the spatial context is lost; (iii) performs effective closed-loop navigation; (iv) benefits substantially from the proposed spatial-context mechanism; and (v) attains state-of-the-art performance together with high runtime efficiency. We first describe the setup, then address each claim in turn.

\subsection{Experimental Setup}
\label{sec:protocol}

\noindent\textbf{Datasets.} We evaluate 8-step trajectory prediction on RECON~\citep{recon}, SACSoN~\citep{sacson}, TartanDrive~\citep{tartandrive}, and SCAND~\citep{scand}, using 100 fixed test queries per dataset shared across methods, following~\cite{nwm}. We further evaluate zero-shot transfer to Go Stanford~\citep{hirose2019deep}. For closed-loop ImageGoal navigation, we use 100 fixed Gibson~\citep{gibson} validation episodes from the known Mosquito scene in Habitat-Sim~\citep{habitat}. These episodes are disjoint from the same-scene trajectories used to fine-tune the navigation model and train the localization interface. We also present a variant of our approach that uses the ground-truth pose to provide spatial context, denoted as \method{-GT}, to show how much our method can benefit from an ideal spatial context.

\noindent\textbf{Baselines.} The trajectory comparison includes direct
navigation policies GNM~\citep{gnm} and NoMaD~\citep{nomad}, predictive world
models NWM~\citep{nwm}, DINO-WM~\citep{dinowm}, DreamerV3~\citep{dreamerv3},
LeWM~\citep{lewm}, and LS-NWM~\citep{lsnwm}, and geometry-aware models
EvoWorld~\citep{evoworld} and Geometry Forcing~\citep{geometryforcing}. Each
comparison uses identical held-out queries or episodes and the same metric
definitions, while methods retain their own prediction and action-selection
procedures. Trainable methods use the corresponding benchmark's training
trajectories for adaptation. Spatial-perceiving WAM~\citep{swam2026}, NavWAM~\citep{navwam2026}, and WAM-Nav~\citep{wamnav2026} are not quantitatively compared because their official implementations were unavailable at the time of submission.

\noindent\textbf{Metrics.} For trajectory prediction, we report absolute
trajectory error (ATE), the mean Euclidean distance between predicted and
ground-truth terminal positions after eight steps, and relative pose error
(RPE), the mean per-step Euclidean error between predicted and ground-truth
waypoint increments. For navigation, we report
For navigation, we report success rate (SR), success weighted by path length
(SPL), and final geodesic distance to goal (DTG). Each episode executes at
least one action, after which the evaluator issues STOP when the geodesic
distance falls below 2~m, following~\citep{lsnwm}. For inference efficiency, we measure single-step and 16-step rollout latency on an H800 using bfloat16 and batch size 16.

% \noindent\textbf{Pose inputs.} \method{} uses R-SCoRe~\citep{rscore} to predict scene coordinates and recover each camera pose from 2D--3D correspondences through PnP-RANSAC. Current, goal, and history frames are localized in the same scene frame. The current and goal poses define the relative-pose condition; consecutive history poses define the causal motion profile. RANSAC inlier statistics provide a confidence measure. 

% \noindent\textbf{Implementation details.} 
% Images are resized to $224\times224$ and encoded into $384$-dimensional DINOv3 patch features. The implicit transition uses eight learned tokens and four Transformer layers, while the joint latent--action decoder uses two additional Transformer layers. We use four context frames and an 8-step horizon. The four-dataset model is trained for $25{,}800$ updates before Gibson fine-tuning. The navigation checkpoint is fine-tuned with a $5\times10^{-6}$ learning rate on a mixture of Gibson expert, long-goal imitation, and recovery trajectories. 

% \FloatBarrier
\subsection{Trajectory Prediction}
\label{sec:trajectory_prediction}
This experiment tests claims (i) and (ii) in the prediction setting.
% : \method{} attains state-of-the-art 8-step trajectory prediction and, with a stronger localizer, retains high potential for even better prediction, an ability that relies on the estimated relative-pose context grounding the rollout in the goal. 
As shown in Table~\ref{tab:prediction}, \method{} quantitatively attains the
best average ATE/RPE of $1.82/0.46$ across the four datasets, substantially
reducing ATE/RPE by $40.1\%/35.2\%$ relative to the second-best baseline. The
improvement is most pronounced on SACSoN and TartanDrive, which we attribute to
the reliable spatial and motion conditions in SACSoN and the diverse motion
scales in TartanDrive. In addition, our method performs comparably to the
baselines on RECON but degrades on SCAND, mainly because cross-platform motion, operator variation, and dynamic pedestrians severely limit localization performance, thereby undermining the gains it provides to the world model.
Note that, when using ground-truth pose, \method{-GT} achieves a substantial
performance gain. This gain indicates that our method holds high potential when
provided with a better spatial context.

\begin{table}[t]
\centering
\caption{8-step trajectory prediction. Bold denotes the best and underlined the second-best value among non-GT methods.}
\label{tab:prediction}
\footnotesize
\renewcommand{\arraystretch}{1.0}
\begin{tabular*}{\textwidth}{@{\extracolsep{\fill}}lccccc@{}}
\toprule
Method
& RECON
& SACSoN
& TartanDrive
& SCAND
& Average \\
\midrule
GNM~\citep{gnm}
& 1.87/0.73
& \secondbest{3.71}/1.00
& 6.65/1.62
& 2.12/1.61
& 3.59/1.24 \\

NoMaD~\citep{nomad}
& 1.95/0.53
& 3.73/0.96
& 6.32/1.31
& 2.24/0.49
& 3.56/0.82 \\

NWM~\citep{nwm}
& \textbf{1.13}/\secondbest{0.35}
& 4.12/0.96
& 5.63/1.18
& 1.28/\secondbest{0.33}
& \secondbest{3.04}/\secondbest{0.71} \\

DINO-WM~\citep{dinowm}
& 2.33/0.54
& 5.39/1.10
& 5.00/\secondbest{1.06}
& 1.87/0.42
& 3.65/0.78 \\

DreamerV3~\citep{dreamerv3}
& 1.88/\textbf{0.34}
& 4.86/1.01
& 5.96/1.24
& 1.38/0.34
& 3.52/0.73 \\

LeWM~\citep{lewm}
& 2.20/0.66
& 4.01/1.05
& \secondbest{4.93}/1.31
& 2.16/0.58
& 3.33/0.90 \\

LS-NWM~\citep{lsnwm}
& 1.60/0.42
& 4.29/0.99
& 5.50/1.28
& \secondbest{1.26}/\secondbest{0.33}
& 3.16/0.76 \\

EvoWorld~\citep{evoworld}
& 3.10/0.72
& 5.86/1.22
& 6.55/1.35
& 2.95/0.63
& 4.61/0.98 \\

Geometry Forcing~\citep{geometryforcing}
& 1.64/0.48
& 4.47/\secondbest{0.94}
& 5.23/1.11
& \textbf{1.15}/\textbf{0.31}
& 3.13/\secondbest{0.71} \\

\midrule
\method{}
& \secondbest{1.38}/0.36
& \textbf{1.18}/\textbf{0.40}
& \textbf{2.70}/\textbf{0.63}
& 2.04/0.45
& \textbf{1.82}/\textbf{0.46} \\

\textcolor{gray}{\method{}-GT}
& \textcolor{gray}{0.46/0.18}
& \textcolor{gray}{0.94/0.35}
& \textcolor{gray}{1.13/0.36}
& \textcolor{gray}{0.57/0.23}
& \textcolor{gray}{0.78/0.28} \\

\bottomrule
\end{tabular*}
\end{table}

\begin{figure}[t]
\centering
\includegraphics[width=1\textwidth]{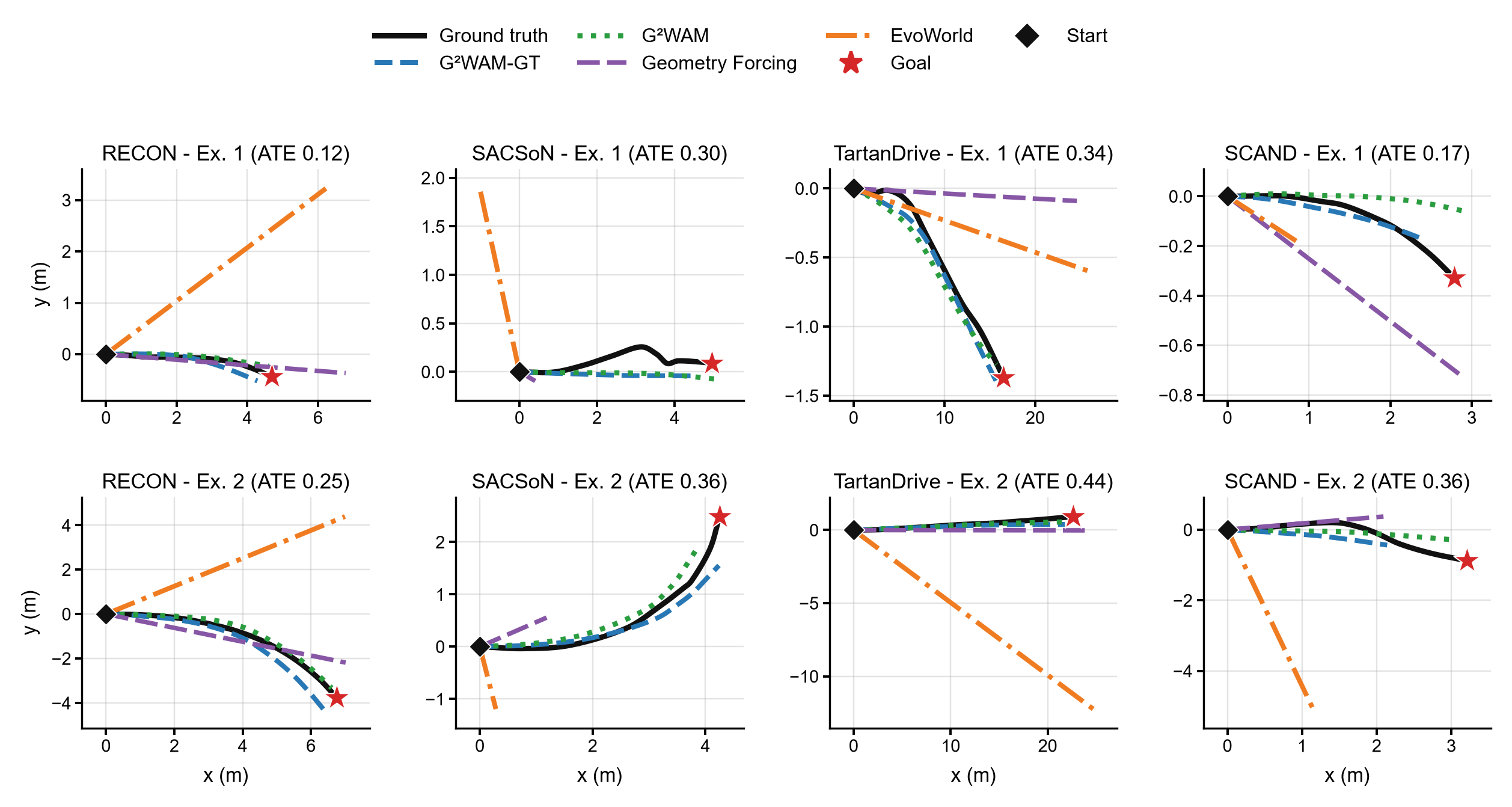}
\vspace{-15pt}
\caption{8-step trajectory rollouts on RECON, SACSoN, TartanDrive, and SCAND, with all paths aligned at the rollout origin. }
\label{fig:traj_examples}
\end{figure}

The qualitative rollouts in Fig.~\ref{fig:traj_examples} illustrate the same
trend. Baseline trajectories often drift away from the target direction or
accumulate scale error over the 8-step horizon, whereas \method{} rollouts
remain better aligned with the ground-truth path. \method{-GT}
yields the cleanest rollouts, and \method{} preserves much of this structure
under an image-based pose interface. These results support the main design
choice of predicting canonical waypoint increments directly from a
goal-conditioned latent transition, rather than relying on a separate sampling
or trajectory-selection stage.

Table~\ref{tab:gostanford} evaluates our method when zero-shot transferred to a new scene without spatial context. Though not the primary scenario of this paper, this setting tests whether our world action model backbone has learned generalizable navigation priors without spatial context. We also report results with a localization module trained on a subset of the dataset, referred to as \method{}-PoseFT, and with ground-truth pose simulating external localization devices such as GPS or UWM. Without spatial context, our method performs comparably to the baselines, indicating that our WAM does not rely solely on spatial context but possesses certain navigation priors. With pose assistance, it improves ATE/RPE by 14.0\%/16.5\%.

% \FloatBarrier
\subsection{Closed-Loop Navigation}
\label{sec:closed_loop_navigation}

\begin{table}[t]
\centering
\begin{minipage}[t]{0.48\linewidth}
\vspace{0pt}
\centering
\captionof{table}{Zero-shot trajectory prediction on Go Stanford.
Bold denotes the best result and underlining the second-best result.}
\label{tab:gostanford}
\footnotesize
\setlength{\tabcolsep}{2pt}
\renewcommand{\arraystretch}{1.05}
\begin{tabular*}{\linewidth}{@{\extracolsep{\fill}}lrr@{}}
\toprule
Method & ATE $\downarrow$ & RPE $\downarrow$ \\
\midrule
NWM~\citep{nwm} & 2.3535 & \secondbest{0.6447} \\
DINO-WM~\citep{dinowm} & 2.8993 & 0.7275 \\
LeWM~\citep{lewm} & 3.0258 & 0.7602 \\
Geometry Forcing~\citep{geometryforcing}
& \textbf{2.0981} & \textbf{0.6063} \\
\midrule
\method{} & \secondbest{2.3096} & 0.6510 \\
\textcolor{gray}{\method{}-PoseFT} & \textcolor{gray}{1.9866} & \textcolor{gray}{0.5434} \\
\textcolor{gray}{\method{}-GT}
& \textcolor{gray}{1.5424} & \textcolor{gray}{0.4296} \\
\bottomrule
\end{tabular*}
\end{minipage}\hfill
\begin{minipage}[t]{0.50\linewidth}
\vspace{0pt}
\centering
\captionof{table}{Gibson image goal navigation. Bold denotes the best result and underlining the second-best result.}
\label{tab:navigation}
\footnotesize
\renewcommand{\arraystretch}{1.05}
\resizebox{\linewidth}{!}{%
\begin{tabular}{@{}lccc@{}}
\toprule
Method & SR $\uparrow$ & SPL $\uparrow$ & DTG $\downarrow$ \\
\midrule
GNM~\citep{gnm} & 24.0\% & 22.8\% & 5.37 \\
NoMaD~\citep{nomad} & 28.0\% & 27.4\% & 4.78 \\
NWM~\citep{nwm} & 32.0\% & 26.35\% & 5.39 \\
DINO-WM~\citep{dinowm}
& \secondbest{54.0\%} & \secondbest{36.4\%} & 4.56 \\
DreamerV3~\citep{dreamerv3} & 44.0\% & 33.7\% & 5.04 \\
LeWM~\citep{lewm} & 42.0\% & 33.01\% & \secondbest{4.51} \\
\midrule
\method{} & \textbf{63.0\%} & \textbf{59.2\%} & \textbf{3.39} \\
\textcolor{gray}{\method{}-GT}
& \textcolor{gray}{71.0\%} & \textcolor{gray}{70.0\%}
& \textcolor{gray}{3.04} \\
\bottomrule
\end{tabular}%
}
\end{minipage}
\end{table}

This experiment tests claim (iii) in the closed-loop navigation. Concretely, it tests whether the predicted waypoints remain useful after their execution changes the observation and the estimated spatial context, rather than only achieving low open-loop trajectory error. Table~\ref{tab:navigation} shows that \method{} achieves 63.0\% SR, 59.2\% SPL, and 3.39~m final DTG. Compared with DINO-WM, the strongest baseline, it improves the two metrics by 9.0 and 22.8 percentage points, respectively. Its final DTG is also 25\% lower than that of LeWM, the strongest baseline on this metric. The improvements across all three measures show that the waypoint predictions remain effective when executed in a repeated observe-and-replan loop, not only in open-loop trajectory prediction. The larger SPL gain over the baselines indicates that, aided by spatial context, our method navigates more purposefully and thus more efficiently. \method{}-GT reaches 71.0\% SR, 70.0\% SPL, and 3.04~m final DTG. The gaps to \method{} indicate that our method still leaves considerable room for improvement. Figure~\ref{fig:difficulty} shows the navigation success rate under different initial start-to-goal distances. The results indicate that spatial context plays a substantial role in both short- and long-range navigation. For long-range navigation, our spatial context does not yield a larger gap over the baselines, possibly because longer distances offer more opportunities for complex structured obstacles to appear between the current position and the goal, which makes direct goal-directed guidance insufficient.

% \FloatBarrier
\subsection{Ablations}
\label{sec:ablations}

\begin{table}[t]
\centering
\caption{Conditioning ablation under the 8-step trajectory-prediction
protocol. 
% For pose--motion configurations, the two labels
% denote the pose and motion sources, respectively; Pred. denotes estimated
% conditions, and Pred./Pred. is the deployed \method{} configuration. GT
% pose-only is separately trained without motion conditioning, whereas Pred.
% pose-only and None mask the
% corresponding inputs of the complete checkpoint at inference.
}
\label{tab:ablation}
\footnotesize
\renewcommand{\arraystretch}{1.0}
\begin{tabular*}{\textwidth}{@{\extracolsep{\fill}}lcccccc@{}}
\toprule
\multirow{2}{*}{Dataset}
& \multicolumn{3}{c}{Pose + motion}
& \multicolumn{2}{c}{Pose only}
& \multirow{2}{*}{None} \\
\cmidrule(lr){2-4}\cmidrule(lr){5-6}
& GT+GT
& Pred.+GT
& Pred.+Pred.
& GT
& Pred.
& \\
\midrule
RECON
& 0.46/0.18
& 1.17/0.33
& 1.38/0.36
& 0.77/0.33
& 1.38/0.37
& 2.02/0.58 \\
SACSoN
& 0.94/0.35
& 1.15/0.39
& 1.18/0.40
& 1.42/0.48
& 1.42/0.46
& 2.32/0.62 \\
TartanDrive
& 1.13/0.36
& 1.70/0.48
& 2.70/0.63
& 3.56/0.96
& 4.53/0.93
& 5.64/1.19 \\
SCAND
& 0.57/0.23
& 1.55/0.38
& 2.04/0.45
& 0.78/0.29
& 1.96/0.44
& 1.96/0.47 \\
\midrule
Average
& 0.78/0.28
& 1.39/0.39
& 1.82/0.46
& 1.63/0.52
& 2.32/0.55
& 2.99/0.71 \\
\bottomrule
\end{tabular*}
\end{table}

This section tests claim (iv): the proposed spatial-context mechanism markedly improves navigation. We ablate the spatial conditions to assess whether goal-relative pose and causal motion provide complementary information for trajectory prediction.

We evaluate the method under different conditioning settings, as in Table~\ref{tab:ablation}. The variant without any spatial context is denoted as \textit{None}; the variant that provides only the goal and current-image poses is denoted as \textit{Pose only}; and the variant that additionally provides the goal, current, and history trajectories is denoted as \textit{Pose+motion}. For \textit{Pose+motion}, we test three combinations using either ground-truth or R-SCoRe-estimated poses, where, e.g., \textit{GT+GT} indicates that both pose and motion use ground truth. The results show that both inputs are necessary. With the current state-of-the-art R-SCoRe pipeline, pose and motion yield average gains of 39.1\%/35.2\% on ATE/RPE, respectively, and with ideal pose estimation or external localization these gains can expand to 73.9\%/60.6\%.

\begin{table*}[t]
\centering
\begin{minipage}[t]{0.57\linewidth}
\vspace{0pt}
\centering
\captionof{table}{Runtime analysis on an NVIDIA H800 PCIe GPU.
Step and 16-step are model-side timings.}
\label{tab:efficiency}
\footnotesize
\setlength{\tabcolsep}{1pt}
\renewcommand{\arraystretch}{1.05}
\begin{tabular*}{\linewidth}{@{\extracolsep{\fill}}llrrr@{}}
\toprule
 Type & Method
       & \shortstack{1-step\\(ms) $\downarrow$}
       & \shortstack{16-step\\(ms) $\downarrow$}
       & \shortstack{Action selection\\(s) $\downarrow$} \\
\midrule
\multirow{3}{*}{L-WM}  & DreamerV3        & 9     & \textbf{33} & 0.65 \\
  & LeWM             & \textbf{2} & \secondbest{34} & \textbf{0.36} \\
  & DINO-WM          & 26    & 423   & 14.45 \\
  \midrule
\multirow{2}{*}{I-WM}  & NWM              &705   & 11430 & 10.98 \\
& Geometry Forcing & 23486 & 47227 & 541.23 \\
\midrule
G-WM  & EvoWorld         & 7048  & 91175 & 150.26 \\
\midrule
G-WAM & \method{}        & \secondbest{4}
                 & 47 & \secondbest{0.51} \\
\bottomrule
\end{tabular*}
\end{minipage}\hfill
\begin{minipage}[t]{0.40\linewidth}
\vspace{0pt}
\centering
\includegraphics[width=\linewidth]{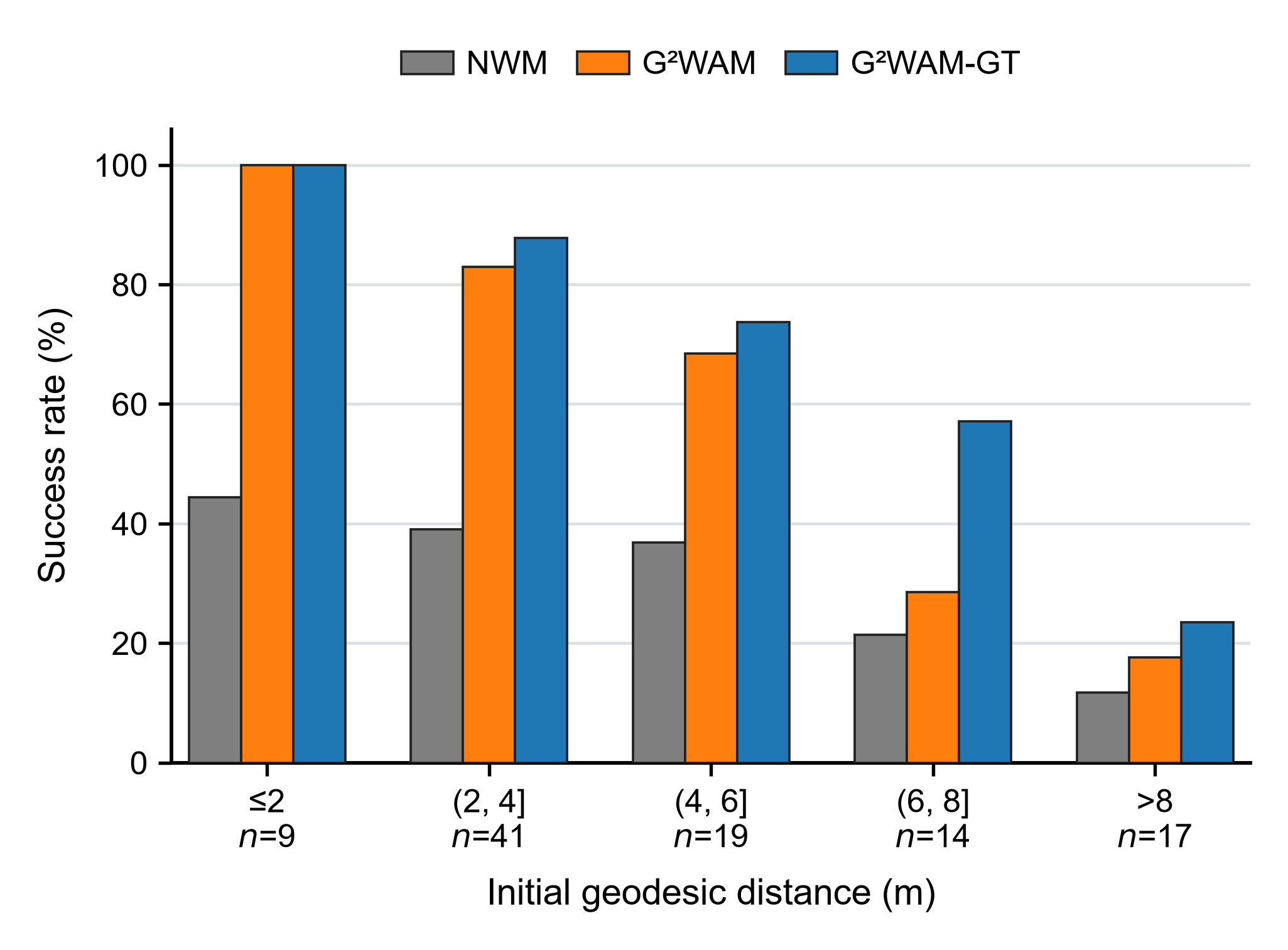}
\vspace{-20pt}
\captionof{figure}{Closed-loop success rate grouped by initial geodesic
distance. \(n\) denotes the number of episodes in each distance range.}
\label{fig:difficulty}
\end{minipage}
\end{table*}

\subsection{Runtime Analysis}
\label{sec:runtime}
% To assess the high computational efficiency of our approach as claimed in (v), we compare inference latency and throughput under a 16-step rollout protocol, as shown in Table~\ref{tab:efficiency}. This measures the model-side cost of predicting latent states and actions from relative-pose spatial context, without pixel-level reconstruction. We report effective FPS as $BL/T_{\mathrm{rollout}}$, where $B$ is the batch size, $L$ the rollout length, and $T_{\mathrm{rollout}}$ the latency in seconds; this measures predicted steps per second, not camera frame rate or full localization-and-control loop frequency. With $B=L=16$, \method{} takes 5.19~ms per step and 54.81~ms for a 16-step rollout, i.e., approximately 4670.5 effective FPS, the lowest single-step and rollout latency among the evaluated methods. The model has $13.5$M trainable parameters plus a 21.6M frozen encoder.

To assess the high computational efficiency of our approach as claimed in (v), we compare inference latency under a 16-step rollout and action-selection protocol, as shown in Table~\ref{tab:efficiency}. We annotate each method with its type: V-WM denotes a video world model, L-WM a latent world model that rolls out in latent space, G-WM a geometry-aware world model, and G-WAM a geometry-aware world-action model. No world action model methods are included, since the relevant works~\cite{swam2026,navwam2026,wamnav2026} are not open-sourced. World-model-based methods generally take longer because they must perform CEM planning. Our method is comparable in efficiency to state-of-the-art L-WM methods and has a clear advantage in action selection. We further analyze the runtime breakdown of action selection. Our method's runtime mainly consists of localization time and world action model inference time, where localization takes 0.42~\!s, which also offers a clear advantage over EvoWorld's geometric-context encoding time of 0.57~\!s. EvoWorld's overall action-selection cost reaches $8.7$~s, since it re-encodes the geometric context at every CEM rollout.

% In Table~\ref{tab:efficiency}, L-WM, I-WM, and G-WM denote latent,
% image-generative, and geometry-aware world models; G-WAM denotes our
% geometry-conditioned world-action model. Step and 16-step report
% model-side latency. Decision includes each baseline's CEM planning
% or \method{}'s online R-SCoRe localization and direct waypoint
% prediction. \method{} takes 3.92~ms per step, 46.68~ms for 16 steps,
% and 0.511~s per decision on the recorded Gibson frame. The methods
% use different prediction workloads and CEM budgets, so these
% timings are not an equal-compute planning comparison.

% \FloatBarrier
\section{Conclusion}
\label{sec:conclusion}

We presented \method, a geometry-grounded world action model for image-goal
navigation. \method{} encodes spatial cues through a learned end-to-end
localization module that anchors the goal and observed frames in a common
reference frame, yielding a stable and consistent spatial context that enables
efficient navigation without explicit 3D reconstruction. We adopt the
world-action-model paradigm and move the rollout into latent space,
jointly predicting future actions and latent future states while avoiding
costly CEM-style planning and pixel-level reconstruction. Extensive experiments
show that our method achieves state-of-the-art trajectory prediction and
navigation performance, all with high inference efficiency.
Future work will focus on extending \method{} to longer-horizon navigation and broader embodied tasks.

\bibliographystyle{iclr2027_conference}
\bibliography{references}
\appendix

\section{Implementation Details}
\label{app:implementation}

\subsection{World-Action Model Architecture}
\label{app:wam_architecture}

\paragraph{Visual encoding.}
All current, historical, and goal images are resized to
$224\times224$ before being passed through a frozen DINOv3
encoder~\citep{dinov3}. The encoder produces patch-level features
with latent dimension $D=384$. We retain the patch structure rather
than spatially pooling the observation features, because the
transition model uses patch-level information to represent spatial
changes across observations.

At planning step $t$, the model receives the four most recent
observation latents,
$\bm z_{t-3:t}\in\mathbb{R}^{4\times P\times D}$, and the goal-image
latent $\bm z_g\in\mathbb{R}^{P\times D}$. The goal latent is first
layer-normalized and then mean-pooled over its $P$ patches. A
two-layer MLP with a GELU activation maps the pooled feature into the
goal embedding:
\begin{equation}
\bm q_g
=
\phi_g
\left(
\operatorname{pool}
\left(
\operatorname{LN}(\bm z_g)
\right)
\right).
\label{eq:app_goal_embedding}
\end{equation}
The DINOv3 encoder remains frozen during world-action-model training,
while the goal projection, transition module, conditioners, joint
decoder, and output heads are optimized.

\paragraph{Relative-pose embedding.}
The relative-pose conditioner receives
\begin{equation}
\bm p_t
=
\left(
\Delta x_t^g,
\Delta y_t^g,
\sin\Delta\theta_t^g,
\cos\Delta\theta_t^g
\right),
\label{eq:app_pose_input}
\end{equation}
where the planar translation is expressed in the current robot frame
and divided by the dataset waypoint spacing. The implementation
separates the translation and yaw components before embedding them.

Let
\begin{equation}
d_t
=
\sqrt{
(\Delta x_t^g)^2+
(\Delta y_t^g)^2
},
\qquad
\bm u_t
=
\frac{
(\Delta x_t^g,\Delta y_t^g)
}{
\max(d_t,\epsilon)
},
\label{eq:app_pose_direction}
\end{equation}
where $\bm u_t$ is the normalized goal direction and $\epsilon$
prevents division by zero. The translation branch receives
$[\bm u_t,\log(1+d_t)]$, while the yaw branch receives
$(\sin\Delta\theta_t^g,\cos\Delta\theta_t^g)$. Each branch is a
two-layer MLP with a GELU activation. Their outputs are concatenated
and passed through another two-layer MLP followed by layer
normalization:
\begin{equation}
\bm e_t^{\mathrm{pose}}
=
\phi_{\mathrm{fuse}}
\left(
\left[
\phi_{\mathrm{trans}}
\left(
\bm u_t,\log(1+d_t)
\right);
\phi_{\mathrm{yaw}}
\left(
\sin\Delta\theta_t^g,
\cos\Delta\theta_t^g
\right)
\right]
\right).
\label{eq:app_pose_embedding}
\end{equation}
This decomposition preserves goal direction, logarithmic distance,
and relative yaw rather than treating the four-dimensional pose
condition as an unstructured vector.

\paragraph{Causal-motion embedding.}
The motion conditioner receives the five-dimensional profile
\begin{equation}
\bm m_t
=
\left[
\bar d_t,
\bar\theta_t,
d_s,
\theta_s,
\ell
\right]^{\mathsf T},
\label{eq:app_motion_input}
\end{equation}
where $\bar d_t$ and $\bar\theta_t$ summarize recent estimated
translation and rotation, $d_s$ and $\theta_s$ are dataset-level
translation and yaw scales, and $\ell$ is the waypoint spacing in
metres. All entries are nonnegative. The implementation applies a
componentwise $\log(1+x)$ transform before passing the profile through
a two-layer MLP with a GELU activation and layer normalization:
\begin{equation}
\bm e_t^{\mathrm{motion}}
=
\phi_{\mathrm{motion}}
\left(
\log(\bm 1+\bm m_t)
\right).
\label{eq:app_motion_embedding}
\end{equation}
The logarithmic transform reduces the effect of scale differences
between datasets and platforms.

\paragraph{Spatiotemporal latent transition.}
The four-frame latent history is first layer-normalized and reshaped
from $\mathbb{R}^{H\times P\times D}$ into a sequence containing
$HP$ patch tokens. Learned temporal and patch positional embeddings
are added to each token:
\begin{equation}
\bm x_{i,p}
=
\operatorname{LN}(\bm z_{i,p})
+
\bm e_i^{\mathrm{time}}
+
\bm e_p^{\mathrm{patch}},
\label{eq:app_position_embedding}
\end{equation}
where $i$ indexes the history frame and $p$ indexes the patch.
Projected goal and relative-pose embeddings are then added to every
spatiotemporal token:
\begin{equation}
\tilde{\bm x}_{i,p}
=
\bm x_{i,p}
+
\phi_{g\rightarrow\mathrm{tok}}(\bm q_g)
+
\phi_{p\rightarrow\mathrm{tok}}
(\bm e_t^{\mathrm{pose}}).
\label{eq:app_transition_conditioning}
\end{equation}

A token-selection module compresses the full spatiotemporal sequence
into eight learned tokens. Concretely, it predicts a normalized
attention map over all $HP$ input patches for each learned token and
uses the resulting attention weights to compute weighted sums of the
input features. The eight selected tokens therefore summarize the
four-frame latent history while retaining information from all
spatial locations.

The goal and relative-pose projections are added again to the learned
tokens before they enter a four-layer Transformer encoder. Each
Transformer layer uses eight attention heads and a feed-forward
dimension of $4D$. The encoded tokens are subsequently fused back into
the original spatiotemporal patch sequence through a learned
token-to-patch mixing module. This operation produces a fused feature
for every patch in every history frame rather than returning only a
small set of global tokens.

The fused patches corresponding to the latest observation are used as
queries in an eight-head cross-attention operation. The complete fused
spatiotemporal sequence serves as keys and values:
\begin{equation}
\bm h_{t+k}
=
\operatorname{MHA}
\left(
\bm x^{\mathrm{last}},
\bm x^{\mathrm{fused}},
\bm x^{\mathrm{fused}}
\right).
\label{eq:app_cross_attention}
\end{equation}
The attended features are combined with the latest-frame patches, a
projected goal bias, and a projected relative-pose bias:
\begin{equation}
\bm c_{t+k}
=
\operatorname{LN}
\left(
\bm h_{t+k}
+
\bm x^{\mathrm{last}}
+
\phi_{g\rightarrow\mathrm{next}}(\bm q_g)
+
\phi_{p\rightarrow\mathrm{next}}
(\bm e_t^{\mathrm{pose}})
\right).
\label{eq:app_patch_context}
\end{equation}
The resulting
$\bm c_{t+k}\in\mathbb{R}^{P\times D}$ is the patch-level transition
context used by both latent-state and action prediction.

The relative-pose embedding is therefore injected at four locations:
the full patch sequence, the learned-token sequence, the
cross-attention residual, and the pooled step state. The
causal-motion embedding is injected only into the pooled step state.
This asymmetry lets the relative pose constrain the spatial evolution
of the latent representation, while recent motion and platform scale
primarily affect action generation.

\paragraph{Joint latent--action decoder.}
The patch-level context is mean-pooled and combined with the pose and
motion embeddings:
\begin{equation}
\bm s_{t+k}
=
\operatorname{pool}(\bm c_{t+k})
+
\bm e_t^{\mathrm{pose}}
+
\bm e_t^{\mathrm{motion}}.
\label{eq:app_step_state}
\end{equation}
This produces a compact
$\bm s_{t+k}\in\mathbb{R}^{D}$ step representation for control.

The joint decoder contains learned action and state-summary query
tokens. The shared step representation is added to both queries.
Patch-level latent tokens are formed by adding a learned latent-type
embedding to $\bm c_{t+k}$. Action, state-summary, and latent-type
embeddings distinguish the three token streams. The resulting token
sequence is processed by a two-layer Transformer encoder with eight
attention heads, pre-normalization, GELU activations, and a
feed-forward dimension of $4D$.

Separate linear heads read the transformed action query and latent
patch tokens. The action head produces the canonical waypoint
increment
\begin{equation}
\hat{\bm a}^{\mathrm{can}}_{t+k}
=
\left(
\Delta x_{t+k},
\Delta y_{t+k},
\Delta\theta_{t+k}
\right),
\label{eq:app_canonical_action}
\end{equation}
while the latent head produces the patchwise candidate
$\bm z^{\mathcal J}_{t+k+1}$. The implementation also retains an
internal state-summary output from the state query, while the
recurrent latent update and executable action are determined by the
latent and action outputs.

\paragraph{Residual latent update.}
In parallel with the joint decoder, a linear base projection produces
\begin{equation}
\bm z^{\mathrm B}_{t+k+1}
=
\phi_{\mathrm B}(\bm c_{t+k}).
\label{eq:app_base_transition}
\end{equation}
The final next-step latent combines the joint-decoder prediction and
the base projection:
\begin{equation}
\hat{\bm z}_{t+k+1}
=
\operatorname{LN}
\left(
\alpha_{\mathcal J}
\bm z^{\mathcal J}_{t+k+1}
+
\alpha_{\mathrm B}
\bm z^{\mathrm B}_{t+k+1}
\right),
\label{eq:app_latent_update}
\end{equation}
where
$\alpha_{\mathcal J}=1.0$ and $\alpha_{\mathrm B}=0.1$.
The base branch provides a direct patch-level transition, while the
joint branch supplies the dominant state-action prediction.

The predicted latent is appended to the fixed-length history and the
oldest latent is removed:
\begin{equation}
\bm z_{t+k-H+2:t+k+1}
=
\left[
\bm z_{t+k-H+2:t+k};
\hat{\bm z}_{t+k+1}
\right].
\label{eq:app_history_update}
\end{equation}
The model repeats this procedure autoregressively for eight rollout
steps. Goal, pose, and motion conditions remain fixed within one
rollout and are recomputed after the next observation in the
closed-loop navigation cycle. No future RGB image is reconstructed
at any rollout step.

\paragraph{Canonical-motion adaptation.}
The joint decoder predicts waypoint increments in a canonical motion
space. The canonical-motion adapter converts these predictions into
the normalized action interface used by each dataset or platform.
The planar components are multiplied by the recent translation scale,
while the yaw component is multiplied by the dataset-level yaw scale:
\begin{equation}
\bm a^{\mathrm{raw}}_{t+k,xy}
=
\tilde d_t
\hat{\bm a}^{\mathrm{can}}_{t+k,xy},
\qquad
a^{\mathrm{raw}}_{\theta,t+k}
=
\tilde\theta_s
\hat a^{\mathrm{can}}_{\theta,t+k},
\label{eq:app_action_scaling}
\end{equation}
where
\begin{equation}
\tilde d_t
=
\operatorname{clip}(\bar d_t,0.1,4.0),
\qquad
\tilde\theta_s
=
\operatorname{clip}(\theta_s,0.05,1.0).
\label{eq:app_action_scale_clipping}
\end{equation}

The scaled planar action is normalized using fixed bounds
$\bm b_{\min}=(-2.5,-4.0)$ and
$\bm b_{\max}=(5.0,4.0)$:
\begin{equation}
\mathcal N_{xy}(\bm\xi)
=
2
\frac{
\bm\xi-\bm b_{\min}
}{
\bm b_{\max}-\bm b_{\min}
}
-\bm 1.
\label{eq:app_action_normalization}
\end{equation}
The final action output is
\begin{equation}
\hat{\bm a}_{t+k}
=
\left[
\mathcal N_{xy}
\left(
\tilde d_t
\hat{\bm a}^{\mathrm{can}}_{t+k,xy}
\right);
\tilde\theta_s
\hat a^{\mathrm{can}}_{\theta,t+k}
\right].
\label{eq:app_action_adapter}
\end{equation}
The relative pose thus supplies target geometry, while the motion
profile determines how a canonical waypoint is converted into
dataset- or platform-specific executable motion.

\subsection{Training Details}
\label{app:training_details}

\paragraph{Training objective.}
The localization interface is trained separately from the
world-action model. The world-action model is optimized using the
multi-term objective in Eq.~\ref{eq:loss}. The latent loss combines
patchwise cosine distance with a smaller mean-squared-error term:
\begin{equation}
\mathcal L_z
=
1-
\operatorname{Cos}
\left(
\hat{\bm Z},
\bm Z^*
\right)
+
0.3
\operatorname{MSE}
\left(
\hat{\bm Z},
\bm Z^*
\right).
\label{eq:app_latent_loss}
\end{equation}
The action loss applies a smooth-$\ell_1$ loss to the canonical
translation and yaw increments. Turning steps receive larger weights
to discourage collapse to an always-forward prediction:
\begin{equation}
\mathcal L_a
=
\frac{
\sum_{b,k}
w_{b,k}
\ell_{\mathrm{SL1}}
\left(
\hat{\bm a}^{\mathrm{can}}_{b,t+k},
\bm a^{\mathrm{can},*}_{b,t+k}
\right)
}{
\sum_{b,k}w_{b,k}
}.
\label{eq:app_action_loss}
\end{equation}
The trajectory and yaw losses compare cumulative executable
translations and rotations over the complete rollout. Planar action
increments are converted back from their normalized representation
before accumulation.

The loss coefficients are
\begin{equation}
\lambda_z=0.3,
\qquad
\lambda_a=3.0,
\qquad
\lambda_{\mathrm{traj}}=2.0,
\qquad
\lambda_{\mathrm{yaw}}=0.5.
\label{eq:app_loss_weights}
\end{equation}

\paragraph{Optimization settings.}
We use AdamW with batch size 32, zero weight decay, and
gradient-norm clipping at 10. The four-dataset checkpoint is trained
for 25,800 parameter updates. It is subsequently adapted to Gibson
using a mixture of expert, long-goal imitation, and recovery
trajectories with a learning rate of $5\times10^{-6}$.

For the reported Mosquito closed-loop experiment, this Gibson-adapted
checkpoint is further fine-tuned for 500 optimization steps on 200
Mosquito trajectories using a learning rate of $2\times10^{-6}$.
These 200 trajectories are disjoint from the fixed 100 evaluation
episodes. They are also used as the reference observations for
constructing the scene-specific localization database; consequently,
the experiment follows a known-scene but held-out-trajectory protocol.
No evaluation episode is used for world-action-model fine-tuning or
localization training.

The DINOv3 encoder remains frozen throughout training and fine-tuning.
The localization interface is trained independently and is not updated
by the world-action-model losses. At inference time, it supplies only
the relative-pose and causal-motion conditions.

\subsection{Localization Interface}
\label{app:localization_interface}

\paragraph{R-SCoRe localization.}
The main text defines $f_{\mathrm{loc}}$ as an interface that maps an
image to a pose in the shared scene frame
(Eq.~\ref{eq:image_to_pose}). We instantiate this interface with
R-SCoRe~\citep{rscore}, a learned scene-coordinate localization
method. R-SCoRe does not directly regress the final camera pose.
Instead, it combines local image descriptors with retrieved global
scene encodings to predict three-dimensional scene coordinates for
query keypoints. A geometric PnP--RANSAC solver then recovers the
camera pose from the resulting 2D--3D correspondences.

The original R-SCoRe evaluation reports a $64.3\%$ localization
success rate within $(0.25\,\mathrm{m},2^\circ)$ and an inference
time of approximately $140$--$270$\,ms per query on an RTX 2080.
These values describe the localization benchmarks and hardware used
in the original R-SCoRe evaluation, rather than the runtime of our
Habitat implementation. The model-side latency reported in the main
paper excludes localization, simulation, and action execution.

\paragraph{Reference-view retrieval.}
For a query image $\bm I_j$, NetVLAD produces a global descriptor and
retrieves a set of candidate reference views from the scene database
$\mathcal D_{\mathrm S}$:
\begin{equation}
\mathcal H_j
=
\operatorname{Retrieve}_{K}
\left(
\operatorname{NetVLAD}(\bm I_j),
\mathcal D_{\mathrm S}
\right).
\label{eq:app_rscore_retrieval}
\end{equation}
The Habitat implementation uses product-quantized nearest-neighbor
search over the stored NetVLAD descriptors. The retrieved view index
$r\in\mathcal H_j$ selects a candidate-specific global scene encoding
$\bm g_r$ from the R-SCoRe reference database.

The NetVLAD descriptor is used only to retrieve candidate views. It
is distinct from the global scene encoding $\bm g_r$ used by the
scene-coordinate prediction network.

\paragraph{Scene-coordinate prediction.}
A local feature encoder extracts query keypoints $\bm u$ and their
descriptors $\bm d_{j,\bm u}$. For each candidate reference view
$r$, its global scene encoding is concatenated with every local query
descriptor. The scene-coordinate head then predicts a three-dimensional
scene coordinate:
\begin{equation}
\hat{\bm X}_{j,\bm u}^{(r)}
=
f_{\mathrm{SCR}}
\left(
[
\bm d_{j,\bm u};
\bm g_r
]
\right).
\label{eq:app_rscore_coordinates}
\end{equation}
Because the global encoding is candidate-dependent, the same query
keypoint produces one scene-coordinate hypothesis for every retrieved
candidate. This allows the localization system to evaluate multiple
possible regions of the known scene.

\paragraph{Pose estimation and candidate selection.}
A separate PnP--RANSAC pose hypothesis is estimated for every
retrieved candidate:
\begin{equation}
\left(
\hat{\bm T}_{C_j\leftarrow\mathrm S}^{(r)},
\mathcal I_j^{(r)}
\right)
=
\operatorname{PnP\!-\!RANSAC}
\left(
\{
(
\bm u,
\hat{\bm X}_{j,\bm u}^{(r)}
)
\}_{\bm u},
\bm K_j
\right),
\label{eq:app_rscore_pnp}
\end{equation}
where $\bm K_j$ is the query-camera intrinsic matrix and
$\mathcal I_j^{(r)}$ is the corresponding RANSAC inlier set. The
selected candidate is
\begin{equation}
r^*
=
\underset{r\in\mathcal H_j}{\arg\max}
\left|
\mathcal I_j^{(r)}
\right|.
\label{eq:app_rscore_candidate}
\end{equation}
Thus, candidate selection is based on geometric consistency rather
than retrieval similarity alone.

The PnP solver returns a scene-to-camera pose. We invert it to obtain
the camera-to-scene pose and then apply the known camera-to-robot
extrinsic calibration:
\begin{equation}
\hat{\bm T}^{\mathrm S}_{B_j}
=
\hat{\bm T}^{\mathrm S}_{C_j}
\left(
\bm T_{B\leftarrow C}
\right)^{-1}.
\label{eq:app_camera_robot_calibration}
\end{equation}
The resulting calibrated robot pose is the
$\hat{\bm T}^{\mathrm S}_j$ used in
Eq.~\ref{eq:image_to_pose}.

\paragraph{Shared-frame spatial context.}
The current, goal, and history images are independently localized in
the same scene frame. Their relative transformations therefore provide
the goal-relative pose and consecutive history motions:
\begin{equation}
\hat{\bm T}_{t\rightarrow g}
=
\left(
\hat{\bm T}^{\mathrm S}_t
\right)^{-1}
\hat{\bm T}^{\mathrm S}_g,
\qquad
\hat{\bm T}_{i\rightarrow i+1}
=
\left(
\hat{\bm T}^{\mathrm S}_i
\right)^{-1}
\hat{\bm T}^{\mathrm S}_{i+1}.
\label{eq:app_relative_transforms}
\end{equation}

These relative transformations are invariant to a common rigid
transformation of the shared scene frame:
\begin{equation}
\left(
\bm G\hat{\bm T}^{\mathrm S}_t
\right)^{-1}
\left(
\bm G\hat{\bm T}^{\mathrm S}_g
\right)
=
\left(
\hat{\bm T}^{\mathrm S}_t
\right)^{-1}
\hat{\bm T}^{\mathrm S}_g,
\label{eq:app_frame_invariance}
\end{equation}
for any $\bm G\in\mathrm{SE}(3)$. The world-action model is therefore
independent of the arbitrary global origin and orientation chosen for
the scene coordinate system. This invariance does not eliminate
localization error: the current, goal, and history images must still
be localized consistently in the shared frame.

The planar translation of
$\hat{\bm T}_{t\rightarrow g}$ is converted to the model coordinate
convention and divided by the waypoint spacing $\ell$ before forming
$\bm p_t$. The consecutive history transformations provide the
history-dependent entries of $\bm m_t$.

\paragraph{Localization modularity.}
The retrieved reference images, NetVLAD descriptors, local keypoint
descriptors, candidate global encodings, predicted scene coordinates,
PnP hypotheses, and RANSAC inlier sets are internal quantities of the
localization interface. They are not provided directly to the
world-action model.

The world-action model consumes only the current-to-goal relative
transformation and the recent causal-motion profile. Consequently,
the localization system can be replaced without changing the
world-action-model architecture, provided that the replacement returns
poses in a consistent metric frame. Possible alternatives include a
traditional place-recognition and pose-estimation pipeline, visual
SLAM, an external tracking system, or another learned relative-pose
estimator.

\paragraph{Known-scene scope.}
R-SCoRe requires a scene-specific reference database and a
scene-coordinate prediction model trained from reference observations.
Our Habitat experiment therefore follows a known-scene protocol. The
fixed evaluation episodes are disjoint from the trajectories used to
construct the localization database and fine-tune the navigation
model, although the reference and evaluation trajectories belong to
the same scene.

This setup evaluates navigation to previously observed places from
held-out trajectories rather than navigation in a completely unseen
environment. Applying the same world-action model in an unseen scene
would require an alternative shared-frame localization source, such
as online mapping, visual SLAM, or a map-free relative-pose estimator.
Once the relative-pose and motion conditions are available, the
world-action-model architecture itself does not need to be changed.

\paragraph{Online localization sequence.}
The goal image is localized once at the beginning of an episode and
its estimated pose is cached. Every newly observed RGB frame is
localized online. Only the current and previously observed images are
used; neither future images nor future actions are available to the
localization interface.

For causal-motion estimation, at most four available history poses
are retained. The pose history is updated only after the corresponding
frame has been observed. The estimated-pose evaluation does not use
ground-truth egomotion, and invalid localization estimates do not
silently fall back to ground-truth poses.

\paragraph{Causal estimated motion.}
Let $\bm r_i$ and $\psi_i$ denote the calibrated robot-frame position
and yaw of history frame $i$. The recent translation scale is computed
as the median nonzero displacement in waypoint units:
\begin{equation}
\bar d_t
=
\operatorname{median}_{i\in\mathcal V_t}
\frac{
\left\|
\bm r_{i+1}-\bm r_i
\right\|_2
}{
\ell
},
\label{eq:app_recent_translation}
\end{equation}
where $\mathcal V_t$ contains the consecutive history pairs with
nonzero translation. The recent rotation scale is the mean absolute
wrapped yaw change:
\begin{equation}
\bar\theta_t
=
\operatorname{mean}_{i}
\left|
\operatorname{atan2}
\left(
\sin(\psi_{i+1}-\psi_i),
\cos(\psi_{i+1}-\psi_i)
\right)
\right|.
\label{eq:app_recent_rotation}
\end{equation}

Before two valid history frames are available, $\bar d_t$ and
$\bar\theta_t$ fall back to their dataset-level translation and yaw
scales. If all available history translations are zero, $\bar d_t$
also falls back to the dataset-level translation scale. Its returned
value is lower-bounded by $0.1$. The remaining motion-profile entries
are the dataset-level translation scale $d_s$, dataset-level yaw scale
$\theta_s$, and waypoint spacing $\ell$.

\paragraph{Habitat coordinate conversion.}
For Habitat, the R-SCoRe pose engine returns a camera-to-world
transformation $\bm T_{WC}$. The current-to-goal camera transformation
is
\begin{equation}
\bm T_{C_tC_g}
=
\bm T_{WC_t}^{-1}
\bm T_{WC_g}.
\label{eq:app_habitat_relative_pose}
\end{equation}
Let $\bm t=(t_x,t_y,t_z)$ denote its translation after applying the
metric conversion factor. Under the Habitat camera convention, the
goal-relative forward and left coordinates are
\begin{equation}
\Delta x_t^g=t_z,
\qquad
\Delta y_t^g=-t_x.
\label{eq:app_habitat_translation}
\end{equation}
To recover relative yaw, the goal camera's forward axis
$(0,0,1)^{\mathsf T}$ is transformed into the current camera frame:
\begin{equation}
\bm f_g
=
\bm R_{C_tC_g}
\begin{bmatrix}
0\\0\\1
\end{bmatrix},
\qquad
\Delta\theta_t^g
=
\operatorname{atan2}
(-f_{g,x},f_{g,z}).
\label{eq:app_habitat_yaw}
\end{equation}
The translation is then divided by the waypoint spacing before forming
the relative-pose condition.

\paragraph{Habitat R-SCoRe settings.}
The input RGB observation is processed at an encoder size whose shorter
side is $480$ pixels. The Habitat camera intrinsics correspond to a
$90^\circ$ horizontal field of view; the focal length used by the pose
solver is therefore set to half the image width, with the principal
point at the image center.

The original R-SCoRe evaluation considers $10$ retrieved candidates,
whereas our Habitat implementation retrieves $20$. Retrieval is
implemented with product-quantized nearest-neighbor search over the
reference-image NetVLAD descriptors. For every retrieved candidate,
its stored global scene encoding is concatenated with the local query
descriptors before scene-coordinate prediction.

We use a maximum PnP--RANSAC reprojection error of $10$ pixels and at
most $10{,}000$ RANSAC iterations. A separate pose is estimated for
every retrieved candidate, and the candidate producing the largest
inlier set is selected.

The R-SCoRe scene coordinates used by the Habitat localization map are
converted to metres using a factor of $100$. The waypoint spacing is
\begin{equation}
\ell=0.2172\,\mathrm{m}.
\label{eq:app_habitat_spacing}
\end{equation}
The fixed dataset-level translation and yaw scales in the Habitat
motion profile are
\begin{equation}
d_s=0.95824505,
\qquad
\theta_s=0.87605652.
\label{eq:app_habitat_motion_scales}
\end{equation}

\paragraph{Localization confidence.}
The RANSAC inlier count is converted into a bounded diagnostic
confidence:
\begin{equation}
c_j
=
\operatorname{clip}
\left(
\frac{
\left|
\mathcal I_j^{(r^*)}
\right|-100
}{
1000-100
},
0,1
\right).
\label{eq:app_rscore_confidence}
\end{equation}
Thus, estimates with at most $100$ inliers receive zero confidence,
while estimates with at least $1000$ inliers receive full confidence.

The confidence and inlier count are recorded for localization
diagnostics. They are not additional obstacle-avoidance, safety, or
navigation-control signals. The relative-pose conditioner described in
the main paper receives the four-dimensional
$(\Delta x,\Delta y,\sin\Delta\theta,\cos\Delta\theta)$ condition.
The numerical localization settings above describe the Habitat
interface and do not necessarily apply to the trajectory-prediction
datasets or the Go Stanford experiment.

\section{Evaluation Protocol and Metric Definitions}
\label{app:protocol}

\paragraph{Trajectory prediction.}
All methods are evaluated on the same fixed 100 test queries per
dataset with an eight-step prediction horizon. Let
$\hat{\bm p}_{n,k}\in\mathbb{R}^{2}$ and
$\bm p^{*}_{n,k}\in\mathbb{R}^{2}$ denote the predicted and reference
planar waypoint positions for query $n$ at rollout step $k$, after
converting all methods to a common waypoint coordinate system. Following
the implementation used to produce the reported results, we compute a
trajectory-level root-mean-square absolute trajectory error and a
one-step relative pose error:
\begin{equation}
\begin{aligned}
\mathrm{ATE}
&=
\frac{1}{N}
\sum_{n=1}^{N}
\sqrt{
\frac{1}{K}
\sum_{k=1}^{K}
\left\|
\hat{\bm p}_{n,k}-\bm p^{*}_{n,k}
\right\|_{2}^{2}
},\\
\mathrm{RPE}
&=
\frac{1}{N}
\sum_{n=1}^{N}
\sqrt{
\frac{1}{K-1}
\sum_{k=1}^{K-1}
\left\|
\left(\hat{\bm p}_{n,k+1}-\hat{\bm p}_{n,k}\right)
-
\left(\bm p^{*}_{n,k+1}-\bm p^{*}_{n,k}\right)
\right\|_{2}^{2}
},
\end{aligned}
\label{eq:app_prediction_metrics}
\end{equation}
where $K=8$. ATE measures the absolute deviation of the complete
predicted trajectory, while RPE measures the error in consecutive
relative translations. Neither metric includes rotational or
image-reprojection error. The four-dataset average is the unweighted
mean over RECON, SACSoN, TartanDrive, and SCAND.

\paragraph{Closed-loop navigation.}
The Gibson evaluation uses 100 fixed validation episodes
from the known Mosquito scene, a $2$\,m geodesic success
threshold, and at most 100 executed actions. At each
planning cycle, \method{} observes an RGB frame, updates
its causal history, estimates goal-relative pose and motion,
predicts eight latent/action steps, executes only the first
waypoint through \texttt{VelocityXZYawMove}, and replans
after the next observation. The evaluated controller does
not decode future RGB frames or add CEM, APF, depth-based
safety control, persistent bypass, or anti-spin logic.
After each executed action, the evaluator uses geodesic distance to trigger
an oracle STOP when the agent is within 2~m. No success check is performed at
reset, so every episode executes at least one action. Geodesic distance is
otherwise not provided as a control input.

Success rate (SR) is the fraction of successful episodes.
Success weighted by path length (SPL) weights each success
by the shortest-path length divided by the larger of that
length and the executed path length. Distance to goal
(DTG) is the mean final geodesic distance over all episodes,
including failures. The model-only latency reported in the
main text excludes localization, simulation, and action
execution. The \method{}-GT comparison uses oracle pose
and causal motion; the deployed \method{} setting uses
image-estimated inputs.

\section{Supplementary Code and Navigation Videos}
\label{app:supplementary}

\paragraph{Code package.}
The anonymous supplementary archive provides the
\method{} model definition, evaluation entry points for
eight-step trajectory prediction and Gibson navigation,
pose and causal-motion interfaces, fixed test-split metadata,
configuration files, and validation scripts. It records
checksums for included files and identifiers for the frozen
assets used in our experiments. Raw datasets and third-party
components, including DINOv3, R-SCoRe, and Habitat-Sim,
are not redistributed. The training implementation and
pretrained checkpoints are withheld during anonymous review
and will be released after acceptance. Accordingly, the
archive documents the evaluation pipeline but is not a
self-contained, weight-inclusive reproduction.

\paragraph{Qualitative navigation videos.}
The ten paired videos show actual first-person RGB observations
rendered by Habitat-Sim after executed actions, rather than
images generated by either world model. Each pair uses the
same Gibson Mosquito episode, starting observation, and
goal image. For \method{}, we replay the original evaluation
actions and verify the geodesic distance-to-goal trace at
every step. NWM is freshly rerun using its six-step
distilled checkpoint because repeated closed-loop inference
with the original checkpoint is impractically slow. NWM is
stochastic, so an individual recording need not match the
outcome of an earlier aggregate evaluation. Playback is
aligned by executed action index rather than wall-clock
inference time; after one side terminates, its final frame
remains visible.

\begin{table}[H]
\centering
\caption{Outcome-selected paired Habitat recordings. S/F denotes
success or failure under the 2\,m threshold. Steps are executed
actions, not video frames. These examples do not estimate full-test-set
SR or SPL.}
\label{tab:appendix_videos}
\small
\begin{tabular}{cccccc}
\toprule
Episode index & Episode ID & \method{} & Steps & NWM & Steps \\
\midrule
92 & 3353 & S & 9   & S & 75  \\
17 & 3403 & S & 21  & F & 100 \\
29 & 3532 & S & 60  & S & 36  \\
19 & 3537 & S & 32  & F & 100 \\
59 & 3310 & S & 12  & S & 1   \\
16 & 3527 & S & 25  & S & 4   \\
64 & 3408 & S & 20  & F & 100 \\
24 & 3557 & S & 46  & F & 100 \\
1  & 3595 & S & 26  & F & 100 \\
83 & 3485 & F & 100 & F & 100 \\
\bottomrule
\end{tabular}
\end{table}

The recordings were deliberately selected for qualitative
inspection: four pairs succeed for both methods, five
succeed only for \method{}, and one fails for both.
This outcome-conditioned subset is not representative
of the complete 100-episode evaluation. Quantitative
comparisons should therefore use the main evaluation
table rather than these selected videos.

\section{AI Use Statement}
Generative AI tools were used for language editing, literature
search, drafting portions of the manuscript, software
implementation, and assisting with the reproduction of
baseline methods. All AI-assisted text, code, references,
and baseline implementations were reviewed and verified
by the authors. The authors take full responsibility for
the final manuscript, methods, results, claims, and artifacts.

\end{document}

%% file: math_commands.tex
\usepackage{amsmath,amsfonts,bm}

\def\eqref#1{equation~\ref{#1}}
\def\1{\bm{1}}

\DeclareMathAlphabet{\mathsfit}{\encodingdefault}{\sfdefault}{m}{sl}
\SetMathAlphabet{\mathsfit}{bold}{\encodingdefault}{\sfdefault}{bx}{n}